# Adaptive catalyst discovery emerges from composable scientific intelligence

Wentao Li⸸[1], Qilong Cai⸸[2], Qi Lei⸸[1], Jinxing Chen[2], Ruixuan Chen[1], Jiangjie Qiu[1], Botian Wang[3], Kai Zhao[1], Honghao Chen[1], Leyi Zhao[1], Zemeng Wang[1], Yijun Li[1], Jun Yin[2], Fangfang Li[5], Hongxuan Liu[4], Yingdong Lu[5], Yawen Ouyang[3], Manu Suvarna[6], Wei-Ying Ma[3], Ya-Qin Zhang[3], Kostya S. Novoselov[7], Hao Zhou*[3,8], Jiong Lu*[2], Xiaonan Wang*[1]

**Affiliations:**

[1] Beijing Key Laboratory of Artificial Intelligence for Advanced Chemical Engineering Materials, State Key Laboratory of Chemical Engineering and Low-Carbon Technology, Department of Chemical Engineering, Tsinghua University, Beijing, China.

[2] Department of Chemistry, National University of Singapore, Singapore, Singapore

[3] Institute for AI Industry Research (AIR), Tsinghua University, Beijing, China

[4] Department of Chemical Engineering, Massachusetts Institute of Technology, Cambridge, MA, USA

[5] State Key Laboratory of Advanced Optical Polymer and Manufacturing Technology, Wanhua Chemical Group Co., Ltd., Yantai 264000, P. R. China

[6] Department of Biotechnology & Enzyme Catalysis, University of Greifswald, Greifswald, Germany

[7] Institute for Functional Intelligent Materials, National University of Singapore, Singapore, Singapore

[8] Shanghai Artificial Intelligence Laboratory, Shanghai, China

*Corresponding authors. Email: zhouhao@air.tsinghua.edu.cn, chmluj@nus.edu.sg, wangxiaonan@tsinghua.edu.cn

⸸ These authors contribute equally.

## Abstract

Catalysis underpins chemical manufacturing, yet knowledge for catalyst design remains dispersed across experimental observations, computational data and catalysis subdisciplines. To overcome this challenge, we present Composable Catalytic Intelligence (CatAI), a physics-grounded modular AI framework that learns 1.8 million structure-property relations and organizes them into 30 physicochemical encoders. This reusable intermediate layer enables task-specific models to guide catalyst discovery for targeted reactions and objectives. Across benchmark tasks, CatAI achieves a 7.8% lower task-averaged prediction error than the next-best models. In experimental validation, CatAI identified a new confined-passable pocket catalyst applicable in industrial homogeneous polymer synthesis with 9 times the activity of the commercial benchmark. In parallel, it guided autonomous closed-loop optimization of a highly active Cu-based geminal-atom photocatalyst for C–O coupling reactions surpassing the traditional Ni-based systems. This approach connects property-level data to catalytic objectives, supporting physicochemical knowledge reuse across mechanistically distinct systems through computational screening and iterative model-experiment feedback.

## Main

Catalyst discovery requires navigating an immense and combinatorial chemical space defined by compositions, structures, reaction conditions, and external stimuli[1–5]. This search is further complicated by the local nature of catalytic function, as changes in composition or structure must be translated into active-site coordination, electronic structure, and physicochemical interactions that govern reactivity[6–9]. This nested complexity makes exhaustive exploration impractical. Experimental synthesis and characterization are labor-intensive, and first-principles methods such as density functional theory (DFT) remain too computationally demanding for broad screening[10–12].

Machine learning (ML) has emerged as a route to address these limitations[13], enabling property predictions orders of magnitude faster than DFT[14,15]. Most existing studies train models on hand-crafted descriptors or graph neural network representations to predict catalytic quantities such as adsorption energies and reaction barriers[16–18], requiring case-by-case customization for specific systems. Machine-learning interatomic potentials (MLIPs) have introduced a pretraining-fine-tuning paradigm that improves transferability and accuracy by reusing backbones trained on large DFT datasets[19–23]. More broadly, these efforts reflect the emergence of physics-grounded AI, in which atomistic geometry, symmetry, and structure-property relations constrain learning beyond statistical correlation. In parallel, generative models have been explored to design new catalyst structures by sampling learned chemical spaces, complementing predictive approaches[24–26].

Despite these advances, current models often generalize poorly when applied to distinct catalytic materials and reaction objectives. Descriptor-based approaches often depend on system-specific representations, whereas pretrained MLIPs provide reusable atomic interactions but do not automatically select the physicochemical information required for each downstream task. This limitation arises because different catalytic problems depend on distinct physicochemical principles, for example, adsorption is governed by surface chemistry, whereas redox activity depends on electronic structure. Consequently, current models struggle to select the most relevant features for task-specific objectives[27,28]. Similarly, generative models face complementary limitations, as data scarcity in emerging systems hinders effective pattern learning[25,29]. Moreover, as new systems at the frontier of catalytic research continue to emerge, data-driven models must adapt to new structure-property relations without losing prior knowledge[30]. Existing architectures rarely support this form of extensibility, as sequential fine-tuning leads to the loss of prior knowledge, preventing cumulative knowledge integration[31]. Addressing these challenges requires organizing dispersed physicochemical knowledge into a reusable intermediate layer. It must select task-relevant representations, adapt to limited data, and accommodate new knowledge without retraining existing modules (Fig. 1a).

Here, we introduce Composable Catalytic Intelligence (CatAI), a modular framework that organizes 1.8 million structure-property relations into 30 independently pretrained encoders, forming a physicochemical knowledge layer (Fig. 1b). Limited target data guide the selection and composition of relevant encoders into a single adapted model. New modules can be added without retraining existing components, allowing the knowledge library to expand with emerging tasks. Combined with literature mining, computational structure generation, and an automated catalyst experimental platform (ACEP), CatAI supports two complementary discovery modes: computational screening followed by synthesis, and continuing model-experiment feedback. Across six benchmark datasets covering adsorption, desorption, dissociation, and transfer energies, CatAI reduces prediction error by 7.8% relative to baseline models. In few-shot prediction, CatAI reduces the mean absolute error by 16.5% relative to the next-best models, and in active learning, it achieves higher Top-3 hit rates. Beyond benchmarking, CatAI demonstrates effectiveness in mechanistically distinct discovery tasks (Fig. 1c). In homogeneous catalysis, CatAI enables high-throughput screening of and identification of industrially relevant metallocene catalysts for polyolefin elastomer (POE) synthesis, identifying a new confined-passable pocket zirconocene (**CPP-Zr**) catalyst with an activity of $187.6 \times 10^4$ kg/(mol·h), representing a 9-fold improvement over the commercial benchmark, and reveals a saturated-bridging strategy for geometric control in molecular catalysts. In heterogeneous photocatalysis, integration with ACEP enables closed-loop optimization and discovery of a Cu-based geminal-atom photocatalyst (**Cu GAPC**), achieving high yields in C–O coupling across diverse chloride substrates compared to the traditionally used Ni-based catalysts. Finally, CatAI's modular design supports a community-driven and continuously evolving platform. Together, the strong predictive capabilities and practical applications in catalysis research and industrial implementation highlight CatAI's potential to drive scalable catalyst discovery.

## A Modular Platform for Catalysis Property Prediction

CatAI supports cross-system catalyst property prediction through a library of pretrained modules, each specialized for a distinct physicochemical property class (Supplementary Fig. 1). Together, these modules form a reusable physicochemical knowledge layer linking property data to downstream catalytic objectives. The current library comprises 30 encoders spanning stability, electronic, thermodynamic, surface, mechanical and transport properties, pretrained on 1.8 M structure-property data points. For each task, these encoders represent the target structures in their respective latent spaces. Known target labels guide the estimation of module relevance, yielding weights that minimize the error of the combined predictions. The same weights are then used to average encoder parameters into a single model, which is fine-tuned on the target data. The pretrained module library remains intact for reuse and extension. We evaluated this design through prediction across catalytic tasks, adaptation with limited labels, and the addition of new knowledge modules.

Across diverse catalytic tasks, we first benchmarked CatAI on six datasets covering adsorption, desorption, dissociation, and transfer energies. Specifically, we evaluated CatAI on three global minimum adsorption energy benchmarks[32] (Alloy-GMAE, FG-GMAE and OCD-GMAE), and on reaction energy changes for other elementary catalytic processes in OC20NEB[33], including desorption, dissociation and transfer steps. We compared CatAI against three baseline classes: graph neural networks trained from scratch (CGCNN, SchNet, DimeNet++, Equiformer, eSCN, GemNet-OC, and AdsMT)[32,34–38], a multi-task pretrained model (JMP)[28], and universal interatomic potentials (ORB-v2, ORB-v3)[19,39]. Across these tasks, CatAI consistently outperformed baselines (Fig. 2a), achieving a 7.8% lower task-averaged mean absolute error (MAE) than the second-best model. Complete per-split results are provided in Supplementary Tables 3–8. The gains were observed across adsorption, desorption, dissociation, and transfer energy predictions. CatAI also achieved the lowest mean MAE in out-of-distribution tests on Alloy-GMAE and FG-GMAE (Extended Data Fig. 1a). We also evaluated four domain-specific tasks spanning thermocatalysis[40] (ethylene binding energy), electrocatalysis[41] (*OH and *O adsorption energies), and photocatalysis[42] (electron affinity). Across these regimes, CatAI achieved the lowest average MAE in thermocatalysis and electrocatalysis, while matching highly customized models on photoredox-relevant properties (Supplementary Fig. 9 and Supplementary Tables 23–26). We next extended CatAI to reaction-level enantioselectivity prediction in asymmetric catalysis, which requires jointly modeling reactants, products, catalysts, and solvents. A role-aware attention readout was introduced to distinguish these components. We evaluated signed $\Delta\Delta G^{\ddagger}$ on the asymmetric hydrogenation of olefins (AHO) dataset and a dataset of chiral phosphoric acid (CPA)–catalyzed thiol additions to *N*-acylimines[43]. Among ten methods, CatAI achieved the lowest mean MAE in four settings and the second-lowest in two (Fig. 2b). These results support transferring the physicochemical knowledge layer from individual properties to reaction-level outcomes.

To assess transferability, we evaluated CatAI in label-scarce settings, where DFT or experimental labels are limited. We considered few-shot prediction and active-learning-driven optimization. For few-shot prediction, models were trained on 8–256 samples from Alloy-GMAE and FG-GMAE. Averaged across datasets, CatAI reduced MAE by 16.5% compared with the next-best models (Fig. 2c,d; Supplementary Section 2.3). Active learning targeted candidates with minimal $*CH_2$ and *O adsorption energies. CatAI achieved Top-3 hit rates of 0.83 for $*CH_2$ and 1.00 for *O (Fig. 2e). These rates measure recovery of at least one of the three lowest-energy candidates. CatAI also reached lower final minima for both adsorbates. Full metric definitions and comparisons are provided in Supplementary Section 2.4 and Figs. 6–8. These results indicate that CatAI can prioritize high-performing candidates under constrained evaluation budgets.

We next examined whether CatAI's improvements arise from task-specific module fusion rather than simple assembly. We compared its default fusion strategy with random equal-weight fusion, single-source transfer, and direct initialization from the jointly pretrained JMP model[28] (Fig. 2f). CatAI achieved the lowest errors across all six datasets, whereas random fusion consistently degraded performance and single-source transfer produced less reliable results across tasks. Control configurations and complete results are provided in Supplementary Section 2.7 and Tables 27–32. Removing highly weighted modules yielded higher mean MAE than removing low-weight modules across three benchmarks (Extended Data Fig. 1b). Together, these comparisons support the selective combination of complementary physicochemical knowledge. Separate analyses of representation uniformity are reported in Supplementary Section 2.6 and Fig. 11.

We next integrate CatAI with experiments through computational screening for homogeneous polymerization catalysts and iterative model–experiment feedback for heterogeneous photocatalytic C–O coupling.

**Large-scale virtual screening for homogeneous catalysis**

We first focus on POE synthesis by ethylene–1-octene copolymerization to produce high-value materials. Metallocene catalyst performance is governed by a multistep network involving competing chain-initiation, monomer-insertion and termination pathways (Supplementary Fig. 13). Catalyst ligand structure influences multiple steps simultaneously, and overall activity emerges from their coupled balance rather than a single rate-limiting step, complicating descriptor-based screening. To address this challenge, we combined mechanistic activity calculations, automated structure generation and CatAI prediction to prioritize candidates for experimental comparison with a commercial POE catalyst (Fig. 3a). The calculations supplied activity labels for task adaptation, whereas the structure generator supplied candidate geometries for CatAI to rank.

We first constructed a curated dataset of reported catalytic activities from the literature (Supplementary Fig. 12), which showed that Group IV metallocenes, particularly zirconium-based systems, consistently exhibit the highest activity, motivating their selection as the primary design space. Within this space, we coupled DFT calculations with kinetic Monte Carlo (KMC) simulations to generate activity labels for 268 catalysts (Fig. 3b). Because this workflow is computationally demanding, it was restricted to this labeled set. DFT-derived rate constants allowed KMC to translate competing polymerization pathways into overall activity. These structure–activity pairs were used to compose and fine-tune CatAI for the polymerization objective. As a surrogate for DFT–KMC, the adapted model ranked 7,097 automatically generated structures by predicted activity. Joint projection of the training catalysts and the screening library revealed a continuous activity landscape (Fig. 3c), consistent with CatAI learning a coherent representation of the broader design space from only 268 catalyst-activity pairs. CatAI achieved an MAE of 7.77 for activity prediction on the full dataset (Fig. 3d), supporting the use of its predicted scores for prioritization.

Using these predictions, we selected three top-ranked catalysts for synthesis and experimental comparison with the commercial POE benchmark (Fig. 3b). **Cat 1** and **Cat 2** were prepared following reported synthetic routes[44,45] while the third candidate, a previously unreported fluorenyl–cyclopentadienyl zirconocene with a saturated ansa bridge, was designated confined-passable pocket zirconocene (**CPP-Zr**). All catalysts were tested in ethylene–octene copolymerization on an automated polymerization evaluation platform (Extended Data Fig. 2a). Because POE production depends on both polymer yield and comonomer incorporation, we evaluated catalytic activity alongside 1-octene incorporation. Catalysts that deliver both high activity and substantial incorporation can efficiently produce elastomeric POE with the short-chain branching essential for flexibility and low-temperature performance. Notably, all CatAI-recommended catalysts outperformed the constrained-geometry titanium catalyst (**CG-Ti**) (Activity = $19.2 \times 10^4$ kg/(mol·h)) (Fig. 3e), a commercial benchmark for POE synthesis. Among these, **CPP-Zr** delivered the highest activity, $187.6 \times 10^4$ kg/(mol·h), and also exceeded previously reported Zr-based metallocenes included in our comparison[46,47] (Supplementary Table 34). Its 1-octene incorporation reached 41.7%, the highest among the tested catalysts, compared with 29.4% for **CG-Ti**. This dual enhancement shows that higher catalytic activity can be achieved while maintaining substantial incorporation of the bulkier comonomer. Furthermore, we validated scalability through 2 L reactor trials under industrially relevant conditions. **CPP-Zr** maintained high catalytic activity at this scale (Supplementary Table 36), supporting practical viability for POE production. Additional polymerization metrics, including molecular weight and dispersity, are reported in Extended Data Fig. 2b,c.

We attribute this performance to the coupling of geometric pre-organization and selective steric control. The fluorenyl fragment and ansa linkage rigidify the active-site environment, reducing conformational search and stabilizing insertion-competent arrangements, while tert-butyl substitution imposes steric pressure that disproportionately disfavors chain-transfer and termination channels without blocking monomer approach. Steric analysis further suggests that the superior performance of **CPP-Zr** arises from a more favorable balance between directional confinement and accessibility along the olefin-approach channel. Relative to **Cat 1** and **Cat 2**, **CPP-Zr** shows the strongest directional enclosure around coordinated ethylene (Extended Data Fig. 2d) and simultaneously exhibits the largest coordination-induced tightening of the local steric pocket (Extended Data Fig. 2e), consistent with a more strongly pre-organized insertion environment. This enhanced confinement does not progress to the overrestricted geometry observed for **CG-Ti**, which combines appreciable directional enclosure with a near-collapsed bottleneck and a substantially reduced passable channel volume (Extended Data Fig. 2f,g). **CPP-Zr** thus occupies an optimal steric regime (Extended Data Fig. 2h,i) that balances organization for productive insertion with sufficient accessibility for monomer approach. This balance is consistent with its higher propagation efficiency relative to the overly open **Cat 1**/**Cat 2** or overconstricted **CG-Ti**.

To explore new ligand designs, we combined large language model (LLM) generation starting from an existing catalyst with a fixed CatAI predictor. High-scoring candidates guided subsequent rounds, yielding 15 prioritized candidates after three rounds. Subsequent DFT calculations supported improved catalytic performance for two third-round candidates, $\mathbf{R3_{007}}$ and $\mathbf{R3_{008}}$ (Supplementary Section 3.7 and Supplementary Figs. 22–25), illustrating how generation and physicochemical knowledge-based prediction can guide new candidate discovery.

The discovery of **CPP-Zr** supports an alternative design principle. Whereas conventional metallocene optimization often focuses on local steric tuning, **CPP-Zr** suggests that high activity can arise from global topological reorganization through saturated bridging. The result illustrates how pretrained physicochemical knowledge can guide experimental discovery and inform catalyst design.

### Closed-loop optimization for heterogeneous photocatalysis

We next deployed CatAI in the complementary mode of closed-loop optimization, targeting visible-light-driven C–O coupling, a pharmaceutically relevant transformation. Dual photoredox catalysis combines photoabsorbers and transition metals, but reliance on homogeneous systems and critical raw materials, including iridium-based complexes, presents sustainability and scale-up challenges[48]. Vilé and co-workers reported heterogeneous Ni single-atom carbon nitride catalysts for coupling of carboxylic acids with alkyl halides, with the reported scope mainly covering alkyl carboxylic acids and bromides[49,50] (Fig. 4a). These limitations motivate exploration of more accessible heterogeneous catalysts, although their discovery remains challenging because of the size of the chemical design space. We therefore integrated CatAI with ACEP in an autonomous, closed-loop workflow to explore heterogeneous photocatalysts. We focused on industrially relevant chloride substrates, which are intrinsically challenging owing to the need to activate strong C–Cl bonds, as a stringent test for data-driven catalyst discovery. This workflow identified **Cu GAPC** for C–O coupling across allylic, propargylic and alkyl chlorides, extending the explored chemistry beyond the established Ni-based systems (Fig. 4b).

To define a tractable design space, we conducted two-stage, AI-assisted literature mining (Fig. 4c). We expanded beyond the limited photocatalytic C–O coupling literature to extract metal–support priors from broader cross-coupling studies (Supplementary Section 4.1). These distributions guided an active-learning library spanning Ni, Cu and Pd centers on heterogeneous supports. The library therefore incorporated literature-derived compositional priors, whereas measured yields supplied the task-specific training labels. All candidate structures are visualized in Extended Data Fig. 3a.

Within this literature-informed space, initial ACEP yield measurements (Fig. 4d) were used to compose and adapt CatAI for heterogeneous photocatalyst optimization. ACEP integrated robotic liquid handling, parallel reactor modules with temperature and light control, and NMR/GC analytics for calibrated yield assessment. The model combined catalyst representations with reaction conditions to predict yields for 3,960 catalyst–condition combinations. Each round selected candidates with high predicted yields and high uncertainty for experimental testing, balancing exploitation and exploration. ACEP executed the selected experiments and quantified product yields under the chosen conditions. These measurements were added to the cumulative training set. The updated model then supplied predictions and uncertainties for the next batch. The acquisition scheme and condition-encoded model are detailed in Supplementary Fig. 27 and Supplementary Section 4.2.

The campaign trajectory, visualized by latent-space projection (Fig. 4e), captures how the closed-loop search evolved. We began with a small, manually assembled initialization designed to span representative metal–support motifs, which established Ni as the best-performing family at the outset. Guided by these early measurements and using C–O coupling between *N*-Boc-*L*-proline (**1**) and cinnamyl chloride (**2**) as the target reaction, Round 0 concentrated on Ni-rich candidates and reached a plateau of 48% yield, whereas Round 1 reweighted the search toward a locally improved Pd-based subset that emerged from the first update and plateaued at 44% yield. In Round 2, the acquisition strategy shifted toward a sparsely sampled Cu region highlighted by the model, increasing yield to 68% and establishing an alternative motif for continued optimization. In Round 3, CatAI used this Cu lead as a seed and concentrated sampling within the Cu-support subspace while continuing to probe uncertain neighborhoods to avoid premature convergence, ultimately identifying a **Cu GAPC** that delivered 90% yield.

Kernel density estimates show how predicted yields changed during the search (Fig. 4f). As iterations progressed, the predictive distribution stabilized in shape, while variance in the high-yield region decreased monotonically. From Round 1 onward, Cu-based catalysts dominated the top 1% of predicted yields, supporting the subsequent focus on Cu. Together with the experimental trajectory, these diagnostics track the model's evolving search strategy. Notably, the optimized catalyst–condition combination was reached after experimentally evaluating only about 4% of the candidate pool, demonstrating sample-efficient discovery under challenging C–Cl activation conditions.

Accordingly, the CatAI model identified the optimal conditions for **Cu GAPC** in the catalytic C–O coupling of *N*-Boc-*L*-proline (**1**) and cinnamyl chloride (**2**). The reaction proceeded smoothly with DMSO as the solvent, 10 mol% **Cu GAPC** as the catalyst, and 2-(*tert*-butyl)-1,1,3,3-tetramethylguanidine (BTMG) as the base. To further test whether the outcome reflected a broadly useful reaction setting rather than a single-substrate solution, we evaluated representative chlorinated substrates (Fig. 4g). We also found that the reaction was not restricted to allyl chlorides. The propargyl chloride substrate **4** was efficiently converted to the desired C–O coupling product **5** with 85% yield. Furthermore, alkyl chlorides such as benzyl chloride and phenylpropyl chloride were compatible, leading to the corresponding products **7** and **9** in 92% and 84% yields, respectively. Control experiments, including comparisons with representative heterogeneous catalysts, demonstrate that the reaction critically depends on both light irradiation and **Cu GAPC**, which together significantly enhance the reaction rate (Supplementary Fig. 29). We therefore further characterized the prepared catalyst to elucidate the structural features of the Cu sites (Extended Data Fig. 3b–g). The proposed geminal Cu arrangement may provide spatially proximal and electronically accessible sites for cooperative substrate activation, while photoexcitation of the PCN framework could facilitate charge transfer to the Cu centers, potentially promoting C–Cl bond cleavage during catalysis.

Together, these results demonstrate CatAI's capacity to navigate data-scarce, mechanistically complex design spaces through closed-loop optimization, enabling the discovery of an unanticipated Cu-based regime that transcends precedent-driven search and advances dechlorinative C–O coupling toward more accessible and scalable synthesis.

**CatAI as a Composable Knowledge Infrastructure in Catalysis Discovery**

We next examined knowledge reuse and extension, followed by the practical value of the catalysts identified with CatAI. At the core of knowledge reuse is task-specific module fusion. CatAI assigns weights according to how well pretrained representations align with the target task. In this way, the model focuses on a small number of relevant physicochemical features rather than learning everything from scratch for each dataset. Across benchmarks and catalytic applications, the selected modules align with known chemical principles (Fig. 5a). For example, adsorption-focused problems, including Alloy-GMAE and electrocatalytic *O and *OH binding, consistently place higher weight on surface-related modules. The POE activity task preferentially favors modules encoding thermodynamic descriptors of homogeneous catalysis, whereas heterogeneous photocatalyst optimization favors metal-center charge descriptors. These task-specific patterns suggest that CatAI selects modules associated with mechanistically relevant knowledge, providing an interpretable basis for cross-task transfer.

Task-specific fine-tuning leaves the pretrained module library intact, allowing new encoders to be added without retraining existing modules. Expanding the library improved predictive performance, with the largest gains observed when added modules aligned closely with the target task (Fig. 5b). Post hoc analysis of test-selected module orders showed budget-dependent gains from composition (Extended Data Fig. 1c). CatAI also showed limited forgetting under a shared task-head-refitting protocol (Extended Data Fig. 1d). These results support extending the library to additional catalytic systems and descriptors. To support community use, we provide an interface for inference and module contribution, with the workflow described in Supplementary Section 5.3 and Fig. 33.

This community-facing workflow links model reuse to decisions that translate into manufacturable materials and application-relevant performance. The CatAI-designed **CPP-Zr** produces lower-molecular-weight POE with improved melt flow and rapid setting, supporting hot-melt adhesive applications (Fig. 5c). Life-cycle analysis reveals a clear environmental advantage (Fig. 5d). Cradle-to-gate carbon emissions for both POE products were evaluated using EchoChemAI[51], based on IPCC 2021 $GWP_{100}$ methodology[52]. Metal substitution reduces the catalyst synthesis footprint from 5.43 to 2.67 kg $CO_2$-eq $kg^{-1}$ (-50.75%), while improved catalytic activity decreases 1-octene consumption, lowering the overall POE process footprint from 10.99 to 9.80 kg $CO_2$-eq $kg^{-1}$ (-10.81%).

Beyond these promising results in homogeneous polymerization, we further highlight the synthetic utility of **Cu GAPC** identified *via* CatAI using drug molecules, natural products, and their derivatives as reaction partners. As shown in Fig. 5e, carboxylic acids readily derived from aspirin, ibuprofen, and naproxen performed well, furnishing products **10–12** in 74–92% yields. Additionally, carboxylic acids derived from salicylic acid, an aspirin analog with an OH group, also participated smoothly, delivering product **13** in 69% yield. Encouragingly, the performance of **Cu GAPC** in the C–O coupling reaction was not diminished by highly reactive hydroxyl groups. Beyond C–O bond formation, $Cu_g$/PCN enabled two additional thermocatalytic transformations. Notably, the $Cu_g$/PCN catalytic system enabled a novel decarboxylative cyclization strategy for constructing the imidazolidin-4-one scaffold, affording **16** in 76% yield. The structure of **16** was unambiguously confirmed by single-crystal X-ray diffraction (CCDC 2597417, Supplementary Fig. 28 and Supplementary Table 37). No cyclization product was detected in the absence of $Cu_g$/PCN, highlighting the essential role of the catalyst (Fig. 5f). Furthermore, the $Cu_g$/PCN catalytic system was successfully extended to $C(sp^3)$–$C(sp^2)$ bond formation via cross-coupling with arylboronic acid **17** (Fig. 5g). Product **19** was obtained in 75% yield through selective functionalization of the $C(sp^3)$–Cl bond, while the terminal alkene remained intact. By contrast, the previously reported heterogeneous $MnO_2$-mediated route requires prior preparation of hydrazone **18** followed by a separate pressurized step, affording a 2:1 mixture of regioisomers in 67% combined yield, whereas $Cu_g$/PCN provides direct and regioselective one-step access to **19** in higher yield. Collectively, CatAI establishes an extensible knowledge infrastructure for catalysis. By enabling continual integration of new data and mechanistic insights, it functions as an evolving platform that bridges predictive modeling with real-world applications, ultimately accelerating discovery and delivering tangible societal benefits.

## Discussion

CatAI connects dispersed property-level data to catalytic objectives through an intermediate layer of reusable physicochemical knowledge. The experiments and benchmarks support task-specific selection and composition of this layer when direct labels are limited. This design makes knowledge selection and combination explicit steps in model construction, connecting reusable property representations to more complex catalytic objectives.

Beyond prediction, CatAI was applied to two catalyst discovery studies with distinct mechanisms. In polymerization, computational screening identified a metallocene catalyst with 9-fold higher activity than a commercial benchmark. In photocatalytic C–O coupling, iterative model–experiment feedback identified **Cu GAPC** after evaluating only ~4% of the candidate pool. Both studies share a common framework for defining the design space and adapting physicochemical knowledge to a catalytic objective. This shows that CatAI can serve as a general discovery framework when combined with domain expertise.

Importantly, CatAI can also function as a community-driven platform. New datasets can be used to train additional modules without affecting existing ones, allowing the system to evolve over time. Users can both perform predictions using current modules and contribute new data to expand the model's capabilities. Looking ahead, this modular approach is well suited for complex catalytic systems where data are limited or scattered. Further integration of CatAI with generative models could enable the design of entirely new catalysts, combining data-driven exploration with chemical understanding to discover materials beyond current intuition.

## Methods

### Design of CatAI

CatAI is a modular framework for catalytic property prediction that stores transferable physicochemical knowledge within a library of pretrained encoders and repurposes it for novel tasks *via* parameter-space composition. Each knowledge module is a standalone geometric graph neural network, equivariant to three-dimensional rotations and translations, that processes atomic coordinates and element identities to produce structural representations for property prediction. To prevent heterogeneous supervision signals from collapsing into a single latent space, modules are trained independently on large-scale property objectives and their parameters are preserved as persistent experts in a centralized bank. This design facilitates downstream adaptation by selectively reusing task-relevant expertise rather than retraining from scratch.

Given a downstream dataset $D=\{(x_k,y_k)\}_{k=1}^{m}$ and a bank of $N$ pretrained modules $\{g_j\}_{j=1}^{N}$, CatAI constructs a task-specific initialization through a training-free modular composition step. We first estimate the relevance of each module to $D$ using frozen parameters. Specifically, we generate embeddings for all inputs $x_k$ and apply leave-one-out $K$-nearest-neighbor label propagation to derive proxy predictions $\{\hat{y}_k^j\}_{k=1}^{m}$ for each module $g_j$. We then determine non-negative fusion weights $w=\{w_j\}_{j=1}^{N}$ by minimizing the proxy mean-squared error on $D$ subject to a simplex constraint:

$$\arg\min_{w} E_D(w) = \arg\min_{w} \frac{1}{m}\sum_{k=1}^{m}\left(\sum_{j=1}^{N} w_j \hat{y}_k^j - y_k\right)^2, \quad \text{s.}t. \sum_{j=1}^{N} w_j = 1, w_j \geq 0. \tag{1}$$

Using the optimized weights $w^*$, we synthesize a single initialized model *via* parameter averaging across the module bank. Let $\theta_j$ denote the parameters of module $g_j$; the composed parameters are defined as

$$\theta_D = \sum_{j=1}^{N} w_j^* \theta_j \tag{2}$$

The resulting model, initialized with $\theta_D$, subsequently undergoes fine-tuning with full-parameter updates on D to yield the final task predictor.

### Pretraining of knowledge modules

We curated a comprehensive, multi-level dataset for module pretraining by integrating structure–property data from open repositories with in-house computational data[23,53].

Supervision for crystalline materials and bulk properties was drawn from the Materials Project[54], OQMD[55], JARVIS[56], the Castelli perovskite formation-energy dataset[57] and the Ricci inorganic transport database[58], together with datasets covering heterogeneous catalytic surfaces and transition-metal complexes from tmQM[59]. Additional molecular property data were obtained from CEPDB[60], and data on framework materials from QMOF[61,62]. The data were organized into 30 supervised tasks encompassing electronic properties, thermodynamic stability, and catalysis-relevant descriptors.

Each module was pretrained on a single task using the same backbone architecture coupled with a task-specific regression head. We optimized using AdamW with AMSGrad for 80 epochs with a batch size of 16, a learning rate of $5 \times 10^{-4}$, a weight decay of 0.1, and gradient clipping at 10. The learning rate followed a warmup–cosine schedule with warmup over the first 5 epochs and cosine decay thereafter, and was further reduced on plateau by a factor of 0.1 with a patience of 5 epochs[63].

**Fine-tuning and evaluation on downstream tasks**

We evaluated CatAI on adsorption-energy and reaction-energy prediction benchmarks, together with additional tasks that span distinct catalytic domains. For adsorption thermodynamics, we used three global minimum adsorption energy (GMAE) benchmarks, Alloy-GMAE, FG-GMAE and OCD-GMAE, in which data point corresponds to a unique surface–adsorbate combination. Alloy-GMAE comprised 11,260 combinations and emphasized surface diversity, spanning 1,916 bimetallic alloy surfaces evaluated with 12 small adsorbates containing fewer than five atoms. FG-GMAE comprised 3,308 combinations and emphasized adsorbate diversity, covering 202 functionalized adsorbates across 14 pure metal surfaces. OCD-GMAE, derived from OC20-Dense, comprised 973 surface–adsorbate combinations spanning 967 inorganic surfaces and 74 adsorbates. For other catalytic processes, we evaluated CatAI on OC20NEB, which uses DFT nudged elastic band calculations to estimate reaction energy and includes 258 desorption, 293 dissociation and 376 transfer pathways. Beyond these standardized benchmarks, we incorporated four additional tasks designed to reflect practical catalytic settings spanning thermocatalysis, electrocatalysis and photocatalysis, including ethylene binding energy, OH* and O* adsorption energies, and electron affinity.

For each benchmark, we first constructed a task-specific initialization $\theta_D$ by module fusion. We estimated module relevance by leave-one-out label propagation using the $K$=5 nearest neighbors in embedding space and converted cosine similarities to weights with a temperature $\tau$=0.2. Unless otherwise stated, fusion weights were computed using up to 800 labeled samples from the target task. We set the sparsity regularization coefficient to $\lambda_{\mathrm{reg}}$=0 and imposed a minimum weight threshold of $\epsilon=10^{-6}$ for numerical stability.

All model parameters were then fine-tuned on the benchmark training sets using AdamW with AMSGrad. We used a batch size of 4 and trained for up to 60 epochs with early stopping based on validation MAE, using a patience of 20 epochs. The initial learning rate was $8\times10^{-5}$ and the weight decay was 0.01. We applied a warmup–cosine decay schedule with warmup over the first 5 epochs using a warmup factor of $10^{-1}$, followed by cosine decay over the remaining training. We additionally applied a reduce-on-plateau rule that monitored validation MAE and reduced the learning rate by a factor of 0.1 after 5 epochs without improvement. We reported performance as the average test MAE across five cross-validation folds.

Few-shot performance was evaluated using subsampled datasets with $n \in \{8, 16, 32, 64, 128, 256\}$ labeled examples. For each sample size, we conducted independent training runs and averaged the results over 5 random draws.

For active learning evaluations, we implemented a sequential acquisition protocol to simulate autonomous discovery. Module fusion was performed using a fixed set of 40 samples. The iterative learning cycle then commenced (Round 0) with 20 randomly selected samples. In each subsequent round, Monte Carlo dropout was used to quantify epistemic uncertainty, and 20 new samples were acquired: the 10 candidates with the most favorable predicted properties (exploitation) and the 10 with the highest uncertainty (exploration). This process was iterated for ten rounds. To assess robustness and account for stochasticity in initialization and acquisition, the entire experiment was repeated across 40 independent trials.

For reaction-level selectivity prediction, we coupled task-composed CatAI backbones to an attention readout across AHO and five CPA benchmarks. Each reaction component was represented by three RDKit-generated conformers, and atom embeddings were aggregated across conformers. The readout distinguished components by their chemical roles and combined their features into a reaction-level representation. This allowed information from multiple components to inform selectivity predictions. Evaluation used the fixed train/validation/test partitions specified in the original benchmark publications, with three independent runs using different random seeds. Mean test MAE was the primary metric for comparison with nine baselines.

**Computational modeling and large-scale screening of POE homogeneous catalysts**

To generate a high-fidelity training set for homogeneous POE catalysis, we assembled a library of 268 Zr-based metallocene catalysts using modular ligand templates. For each catalyst, we considered two competing monomer-insertion pathways: ethylene insertion and 1-octene insertion. Each pathway was decomposed into three elementary transformations corresponding to chain initiation, chain propagation and chain termination. For every elementary transformation, we optimized the reactant, transition-state and product structures and extracted the corresponding activation free energy $\Delta G^{\ddagger}$.

All geometries were optimized using DFT as implemented in Gaussian 16 at the B3LYP/def2-SVP level. Harmonic frequency calculations were performed for each stationary point to obtain thermal corrections and to verify stationary-point character, with minima showing no imaginary frequencies and transition states showing a single imaginary mode along the reaction coordinate. Free-energy barriers $\Delta G^{\ddagger}$ for each elementary step were then converted to rate constants using the Eyring equation under consistent stoichiometric conventions, allowing ethylene and 1-octene insertion channels to be compared on a common kinetic basis.

These step-specific rate constants were used to parameterize a KMC model that explicitly simulates the competition between productive insertion events and chain-transfer or termination processes. By running the KMC simulations across relevant operating conditions, we obtained macroscopic activity metrics together with emergent polymer statistics, including $M_n$, $M_w$, dispersity ($Đ$) and comonomer incorporation. Catalytic activity was defined as the mass of polymer produced per mole of catalyst per unit time.

$$\text{Activity}=\frac{m_{\text{polymer}}}{n_{\text{catalyst}} \cdot t}\ \text{kg/(mol·h)} \tag{3}$$

This DFT-to-KMC workflow thus establishes a self-consistent mapping from elementary-step energetics to experimentally interpretable performance descriptors, which serves as the ground truth for training CatAI.

To extend the screening beyond the labeled set, we enumerated 7,097 unlabeled metallocene candidates by systematic permutations of ligand scaffolds and bridging motifs. For each generated catalyst, we constructed 3D starting geometries of reactants using our in-house package ComplexGen, which builds metal–ligand complexes from the ligand SMILES and a predefined metal coordination geometry and then refines ligand conformations by a purely geometric optimization procedure. CatAI was then deployed in a screening mode to score the full library and rank candidates for experimental validation.

**Synthesis of POE homogeneous catalysts**

**CG-Ti**, **Cat 1**, and **Cat 2** were prepared under an inert atmosphere with reference to previously reported methods[44,45,64].

**CPP-Zr** was not previously reported and was also synthesized in two steps. For ligand **L3** synthesis: Under nitrogen atmosphere, 2,7-di-tert-butyl-9H-fluorene (6.96 mmol, 1 equiv) was dissolved in 8 mL MTBE and cooled to 0 °C. 3.8 mL *n*-hexyllithium in hexane (2.2 M, 1.2 equiv) was added dropwise. The mixture was stirred at room temperature for 2 h. Cyclopenta-2,4-dien-1-ylidenecyclohexane (8.35 mmol, 1.2 equiv) was then added and stirred at room temperature overnight. Subsequently, the crude reaction mixture was quenched with EtOH (2 mL). The solution was then washed with 10 mL of saturated $NH_4Cl$ solution. The aqueous phase was extracted with EtOAc (50 mL × 3). The combined organic layers were dried with $Na_2SO_4$, filtered, and concentrated under reduced pressure. Finally, the reaction mixture was purified by silica gel column chromatography to afford the **L3** as a white solid, 652 mg, 22% yield.

For catalyst synthesis: In the glovebox, the **L3** (0.41 mmol, 1 equiv) was dissolved in $^{n}Bu_2O$ (2 mL) and cooled to 0 °C. 0.5 mL *n*-hexyllithium in hexane (2.2 M, 2.7 equiv) was added dropwise and the resultant red solution was heated to 100 °C and stirred for 6 h. After cooling to room temperature, zirconium (IV) chloride (0.45 mmol, 1.1 equiv) was added. The resultant orange solution was stirred for 1 h. At this point, the solution was pumped to dryness, dissolved in toluene and filtered over a medium glass frit packed with Celite to remove LiCl salt. The filtrate was concentrated to 10 mL and placed in a -15 °C freezer to crystallize overnight. This produced 153 mg (56% yield) of red solid which was pure by $^{1}$H-NMR. $^{1}$H NMR (400 MHz, $C_6D_6$) $\delta$ 8.04 (d, $J$ = 8.8 Hz, 2H), 7.73 – 7.61 (m, 4H), 6.33 (t, $J$ = 2.7 Hz, 2H), 5.68 (d, $J$ = 2.7 Hz, 2H), 3.43 – 3.35 (m, 2H), 2.39 – 2.28 (m, 2H), 2.05 – 1.95 (m, 4H), 1.37 (s, 18H), 0.98 – 0.87 (m, 2H) ppm.

**Polymerization evaluation of POE homogeneous catalysts**

The ethylene-1-octene copolymerization performances of all catalysts were evaluated in a parallel pressure reactor following an identical operational procedure, with all operations conducted under an inert atmosphere. All reactants were sequentially introduced into the reactor *via* an automatic syringe pump according to a precisely controlled feeding protocol. Initially, 11 mL of Isopar E solvent and 5 mL of 1-octene monomer were charged into the reaction autoclave, followed by the injection of 100 μL of MMAO-7 as the scavenger and activator. The reactor was then heated to 140 °C as the typical reaction temperature. Upon reaching the preset temperature, ethylene monomer was continuously fed into the sealed reaction system to maintain a constant pressure of approximately 465 PSI throughout the entire polymerization process. The catalytic components were subsequently added stepwise in a specific sequence: 1 mL of a 0.015 mmol/L stock solution of the polymerization catalyst in toluene and 3 mL of Isopar E as the cosolvent. Finally, an appropriate volume of Isopar E solvent was supplemented to keep a constant total solvent volume of 20 mL for each independent reaction system. The duration of the polymerization reaction was 5 minutes, and the reaction was immediately quenched by pressurizing the reactor to 550 PSI with $CO_2$. After the completion of the reaction, the crude polymer product was collected and isolated *via* a standard post-treatment procedure. Catalytic activity was calculated as the mass of dried polymer per mole of catalyst per hour and is reported in units of $10^4$ kg/(mol·h).

The polymer characteristics of the resulting POE samples were determined by Gel Permeation Chromatography (GPC) on a Polymer Char's GPC-IR® equipment with an infrared detector (IR5) in 1,2,4-trichlorobenzene at 150 °C versus polystyrene standards. Number-average molecular weight ($M_n$), weight-average molecular weight ($M_w$), dispersity ($Đ = M_w/M_n$) and 1-octene incorporation were obtained from the GPC-IR traces. The 1-octene incorporation is reported as mol% 1-octene in the copolymer chain.

$$\text{1-octene incorporation} = \frac{n_{\text{octene}}}{n_{\text{ethylene}}+n_{\text{octene}}}\times 100\ \% \quad (4)$$

## Active learning for catalyst–condition optimization in photocatalytic C-O cross coupling

The active-learning campaign targeted heterogeneous photocatalysts for dechlorinative C–O cross-coupling. The search space comprised Ni, Cu and Pd centers immobilized on representative heterogeneous supports, including carbon materials, PCN and metal oxides. Each candidate was defined as a catalyst–condition pair, combining the identity of the metal/support architecture with the reaction settings used in screening.

CatAI served as the acquisition model. To obtain a physically grounded warm start, we first used an initial set of 24 labeled experiments to assign module weights from the CatAI library, thereby identifying the pretrained physicochemical priors most relevant to the target photocatalytic setting. The resulting model was then adapted to the experimental objective, with reaction conditions including base, solvent and irradiation wavelength encoded in a conditional branch and injected *via* feature-wise linear modulation layers to modulate the atomistic representation, allowing the same catalyst scaffold to be evaluated across distinct operating conditions.

In each round, CatAI ranked the full candidate set by predicted yield and estimated epistemic uncertainty using Monte Carlo dropout. Uncertainty was computed from 15 stochastic forward passes with dropout activated, yielding a predictive distribution for each candidate.

Forty catalyst–condition pairs were selected per iteration using a two-stream policy that balances exploitation and exploration. Twenty candidates were chosen by the highest predicted mean yield, and twenty were chosen by the largest predictive standard deviation. Experimental outcomes were appended to the training set, and CatAI was updated on the cumulative data after each round to guide subsequent selection.

## Catalyst synthesis and general experimental setup for metallaphotocatalytic C-O coupling

The preparation of **Cu GAPC** followed a previously reported two-step annealing protocol, $CuCl_2$ (0.36 g) and exfoliated PCN (0.70 g) were dispersed in 200 mL formamide, sonicated for 10 min and then stirred in an oil bath (120 °C) for 12 h, following centrifugation and washing thoroughly by ethanol several times. The oven-dried powder (80 °C) was subsequently heated to 500 °C with a heating rate of 2 °C $min^{-1}$ and kept for 5 h with the protection of $N_2$ flow. Other heterogeneous single-atom catalysts were prepared according to the previously reported protocol[65–67].

**Cu GAPC** (12.6 wt%, 5.0 mg, 10 mol%), *N*-Boc-*L*-proline **1** (0.1 mmol, 23.2 mg), cinnamyl chloride **2** (22.8 mg, 0.15 mmol) and BTMG (34.2 mg, 0.2 mmol), DMSO (0.5 mL) and a magnetic stirrer bar were transferred into a 10 mL vial equipped with a screw-cap rubber septum, which in turn functioned as the batch photoreactor. All compounds were weighed using a high-throughput automated system inside an argon-filled glovebox. Photocatalytic batch experiments were irradiated for 12 h in a PhotoCube™ reactor, and products were analyzed by $^{1}H$ and $^{13}C$ NMR spectroscopy. During system optimization, quantitative data on conversion and product selectivity were obtained by taking a 50 μL aliquot of the reaction mixture with tridecane as an internal standard, diluting with ethyl acetate, filtering, and analyzing by gas chromatography (GC). The C–O coupling products **5**, **7** and **9–13** were prepared using this method or closely related procedures. Conditions for **Cu$_g$/PCN**-catalyzed decarboxylative cyclization yielding **16** and C–C coupling yielding **19** are shown in Fig. 5f,g. Further details on catalyst synthesis and characterization, experimental setups, and the overall metallaphotocatalytic protocol are provided in the Supplementary Information.

## Data availability

The data supporting the findings of this study are publicly available in the Zenodo repository at https://zenodo.org/records/19412454, including the catalyst task data, pretraining data, and modules used in this work. The crystallographic data for product **16** have been deposited at the Cambridge Crystallographic Data Centre under deposition number 2597417. All other data are available in the main text or the supplementary materials.

## Code availability

The source code is available at https://github.com/littlepeachs/CatAI.

## Acknowledgments

National Key R&D Program of China grant 2022ZD0117501 (X.W.)

Scientific Research Innovation Capability Support Project for Young Faculty grant ZYGXQNJSKYCXNLZCXM-E7 (X.W.)

Tsinghua University Initiative Scientific Research Program and China Petroleum & Chemical Corporation grant 223243 (X.W.)

National Research Foundation, Prime Minister's Office, Singapore, under the Competitive Research Program Award grant NRF-CRP29-2022-0004 (J.L.)

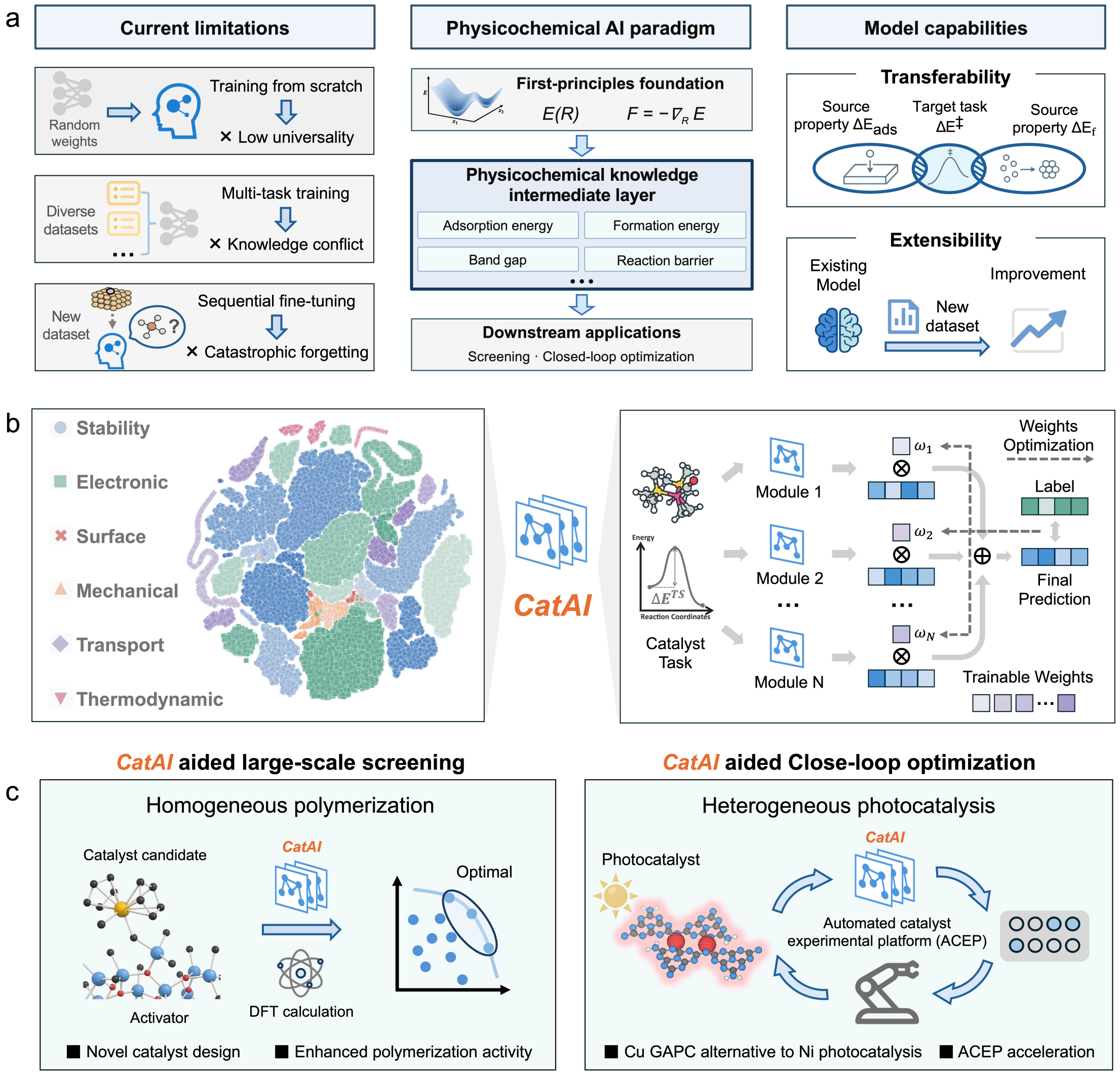

**Fig. 1 | CatAI converts fragmented catalytic knowledge into composable physicochemical priors for catalyst discovery.** (a) Conceptual framework addressing limitations of training from scratch, multi-task learning, and sequential fine-tuning. At its core, an intermediate layer of physicochemical knowledge connects first-principles descriptions of atomic interactions to downstream screening and closed-loop optimization. This layer supports knowledge transfer across properties and extension through additional modules. (b) CatAI's modular architecture. Left: 30 pretrained knowledge modules spanning six physicochemical property families, visualized by t-SNE. Right: task-relevant modules are selected and fused with learnable weights into a unified prediction backbone supporting universality, transferability, and extensibility. (c) CatAI deployed across two mechanistically distinct catalysis domains. In homogeneous polymerization catalysis, CatAI-guided screening identifies a metallocene catalyst **CPP-Zr** for POE synthesis with substantially improved activity. In heterogeneous photocatalysis, CatAI-guided closed-loop optimization identifies a cost-effective **Cu GAPC** for dechlorinative C–O cross-coupling.

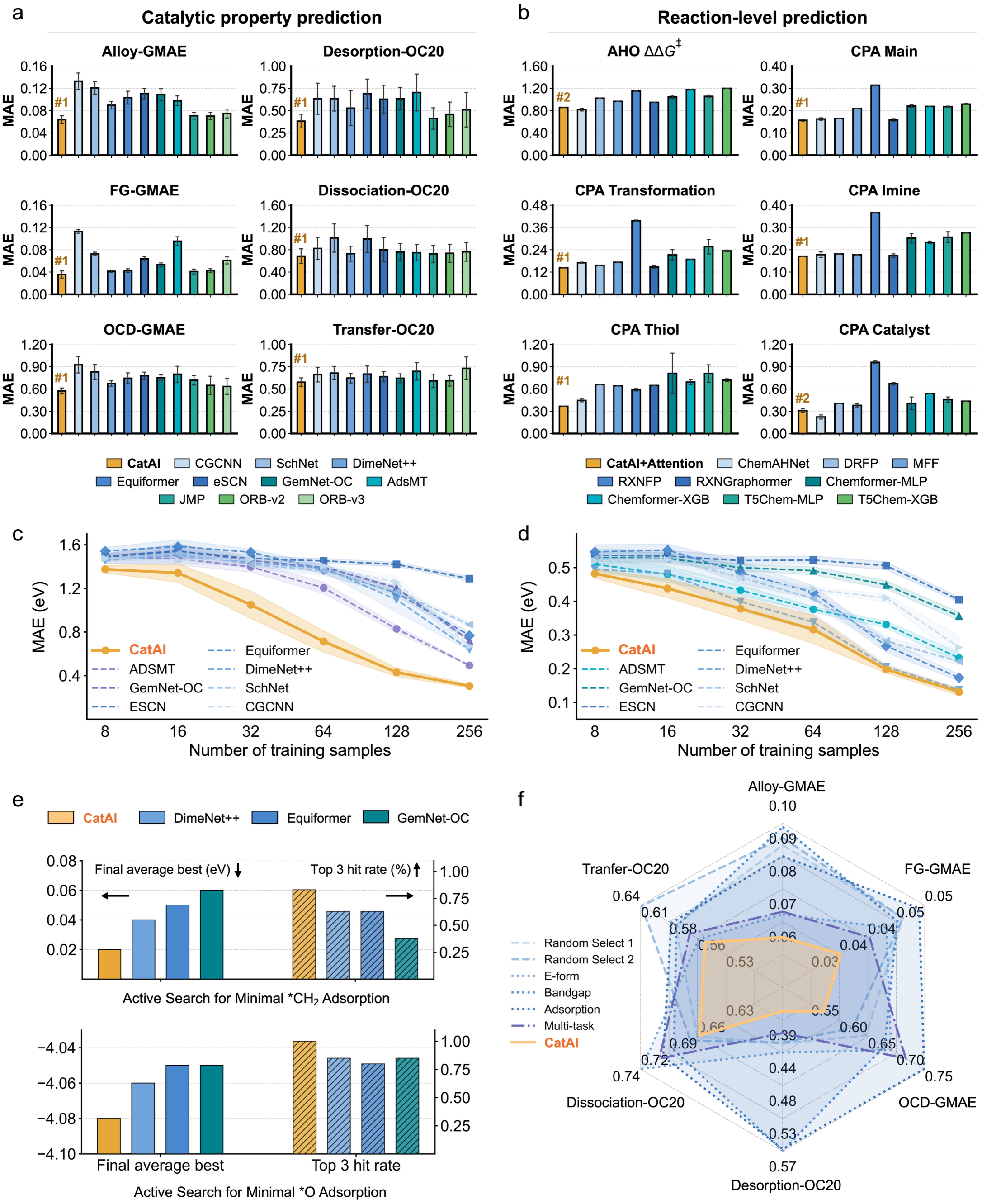

**Fig. 2 | CatAI in silico evaluation across canonical, few-shot and active-learning benchmarks**. (a) Mean absolute error (MAE) on six adsorption and reaction energy benchmarks, evaluated by five-fold cross-validation. (b) Reaction-level prediction on AHO signed $\Delta\Delta G^{\ddagger}$ and five CPA settings, using the original published data splits and three independent runs with different random seeds. This prediction jointly considers reactants, products, catalysts, and solvents. Bars show mean test MAE. (c) Few-shot evaluation under sparse supervision in Alloy-GMAE. Models are trained with 8–256 labeled samples taken from Alloy-GMAE. (d) Few-shot evaluation under sparse supervision in FG-GMAE. (e) Active-learning benchmark for optimizing $*CH_2$ and $*O$ adsorption energies. Upper panels report the final best objective value reached by each method, and lower panels report the Top-3 hit rate achieved during the search. (f) Ablation of module fusion mechanism. Radar plot compares module fusion with random module selection and single-source transfer learning across multiple metrics.

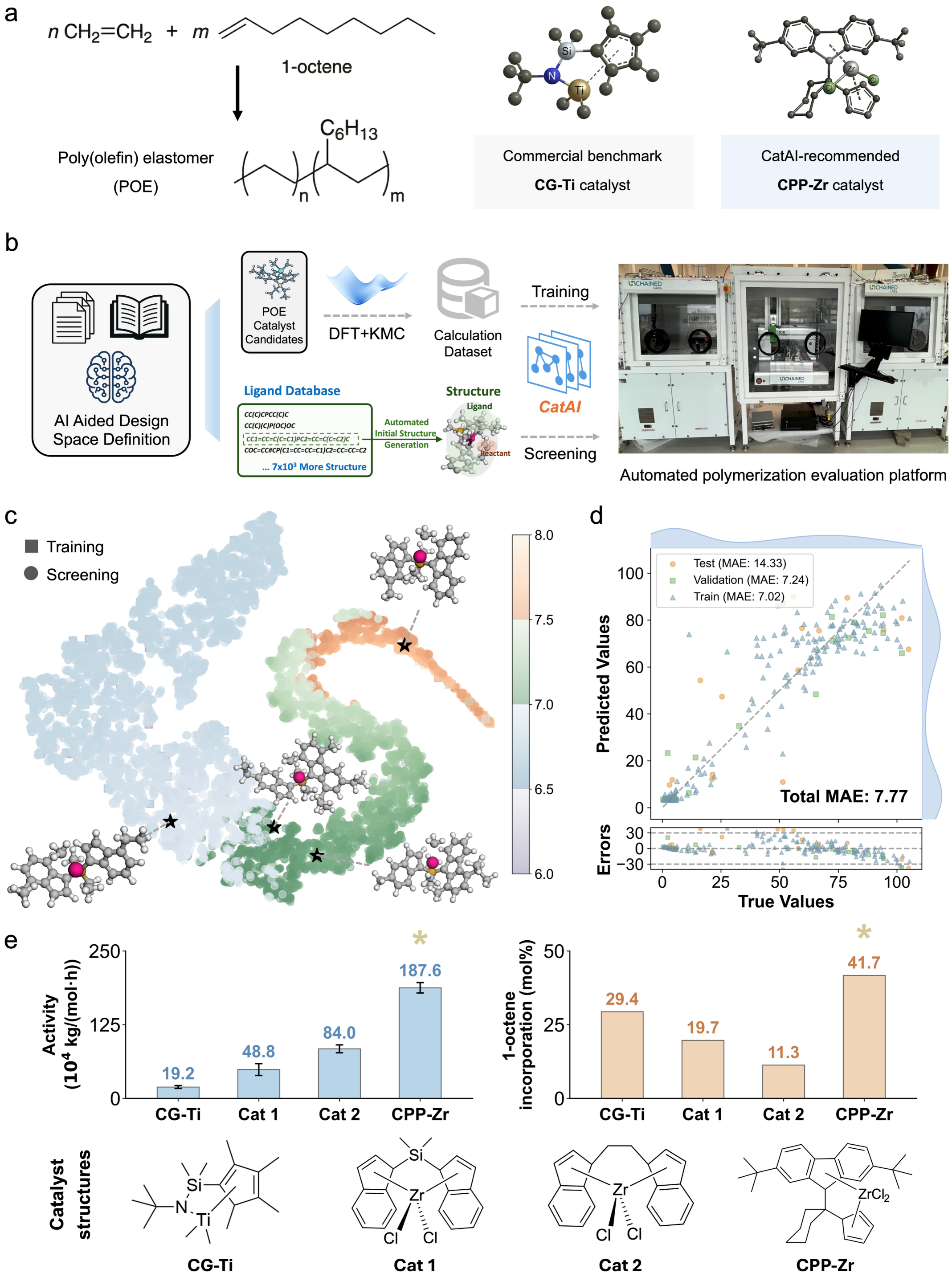

**Fig. 3 | CatAI-driven discovery of homogeneous polymerization catalysts.** (a) Catalytic strategies for ethylene/1-octene copolymerization to produce POE, comparing a commercial benchmark catalyst with a CatAI-recommended catalyst. (b) Discovery workflow. High-throughput DFT and KMC simulations generate the training dataset (left), enabling CatAI to screen 7,097 candidates for validation on an automated polymerization evaluation platform (right). (c) Chemical space projection. t-SNE visualization of the training and screening datasets, colored by CatAI-predicted activity. Black stars highlight representative molecular structures displayed as insets to illustrate chemical diversity. (d) Model accuracy. Parity plot demonstrating high predictive fidelity of catalytic activity (MAE = 7.77 × $10^4$ kg/(mol·h)). (e) Experimental benchmarking of POE catalysts. Catalytic activity and 1-octene incorporation for the commercial benchmark and three recommended catalysts. **CPP-Zr** delivers record activity (187.6 × $10^4$ kg/(mol·h), 9-fold increase) with the highest 1-octene incorporation (41.7%), outperforming the commercial benchmark in both metrics.

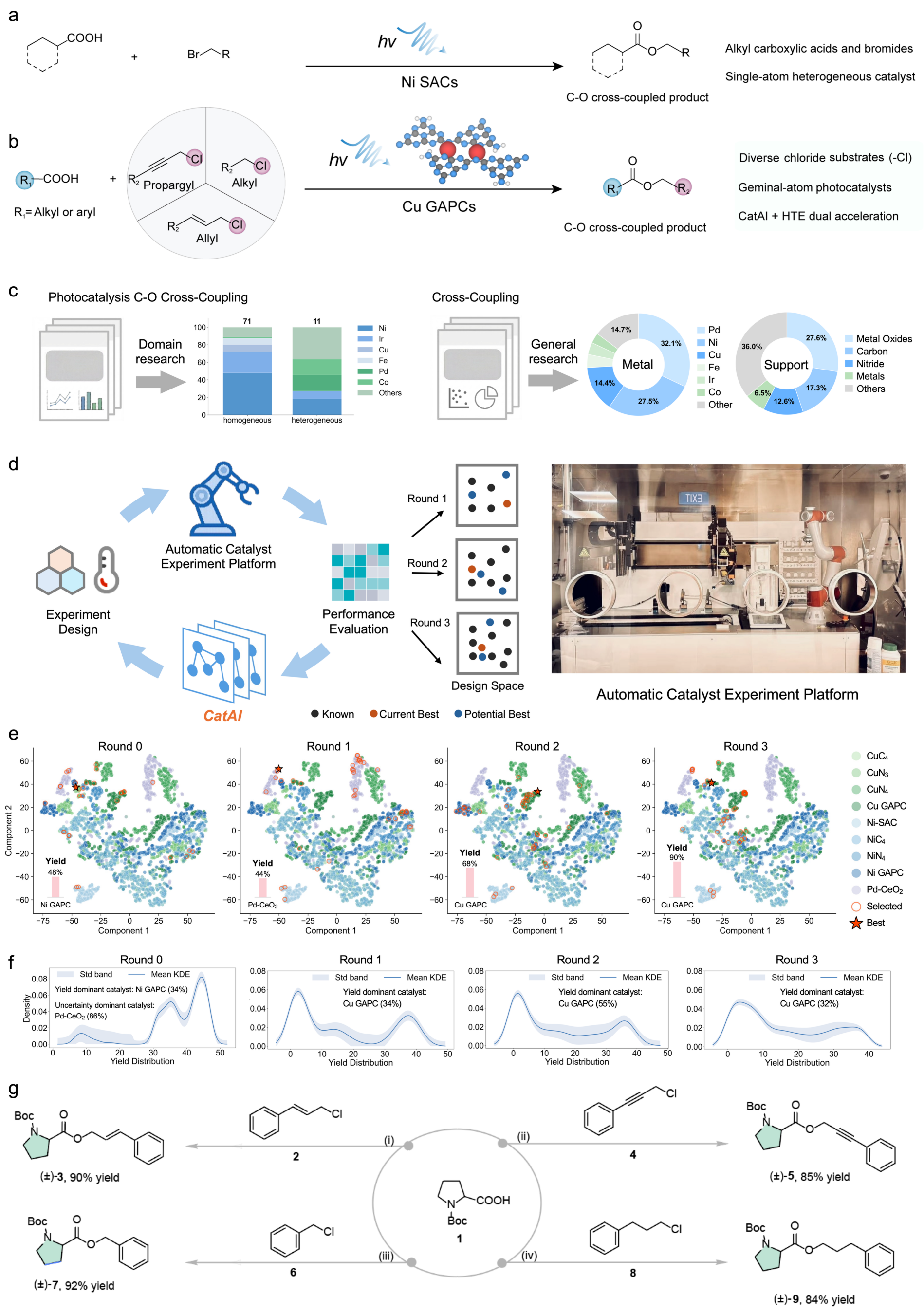

**Fig. 4 | Autonomous catalyst-condition optimization for metallaphotocatalytic C-O coupling.** (a) Established heterogeneous catalytic strategies for C–O cross-coupling. (b) Our heterogeneous catalytic strategies for C–O cross-coupling. (c) Data-driven initialization of design space. Domain research reveals a scarcity of prior art in heterogeneous photocatalytic C–O coupling. To circumvent this limitation, general research was employed to expand the search boundary, identifying Ni, Cu, and Pd motifs anchored on metal-oxide or carbon-nitride supports as the statistically dominant candidates to construct the final design space. (d) Closed-loop active learning. CatAI directs the ACEP to synthesize and test candidates based on predicted high yield and high uncertainty, iteratively updating the model. (e) Evolutionary search trajectory. t-SNE visualization showing the optimization path. The search navigates from suboptimal Ni (48%) and Pd (44%) regions to a high-performance **Cu GAPC** (90%). (f) Kernel density estimates of predicted yield distributions across active learning rounds. (g) Substrate scope under the closed-loop-identified conditions. Representative C–O cross-coupling reactions were carried out with the **Cu GAPC** using the reaction conditions identified by the active-learning campaign. Reaction conditions: *N*-Boc-L-proline **1** (0.1 mmol, 23.2 mg), chloride substrates (0.15 mmol), BTMG (34.2 mg, 0.2 mmol), **Cu GAPC** (10 mol%), DMSO (0.5 mL). Photocatalytic batch experiments were irradiated (420-430 nm) for 12 h using a PhotoCube$^{TM}$ reactor, isolated yields.

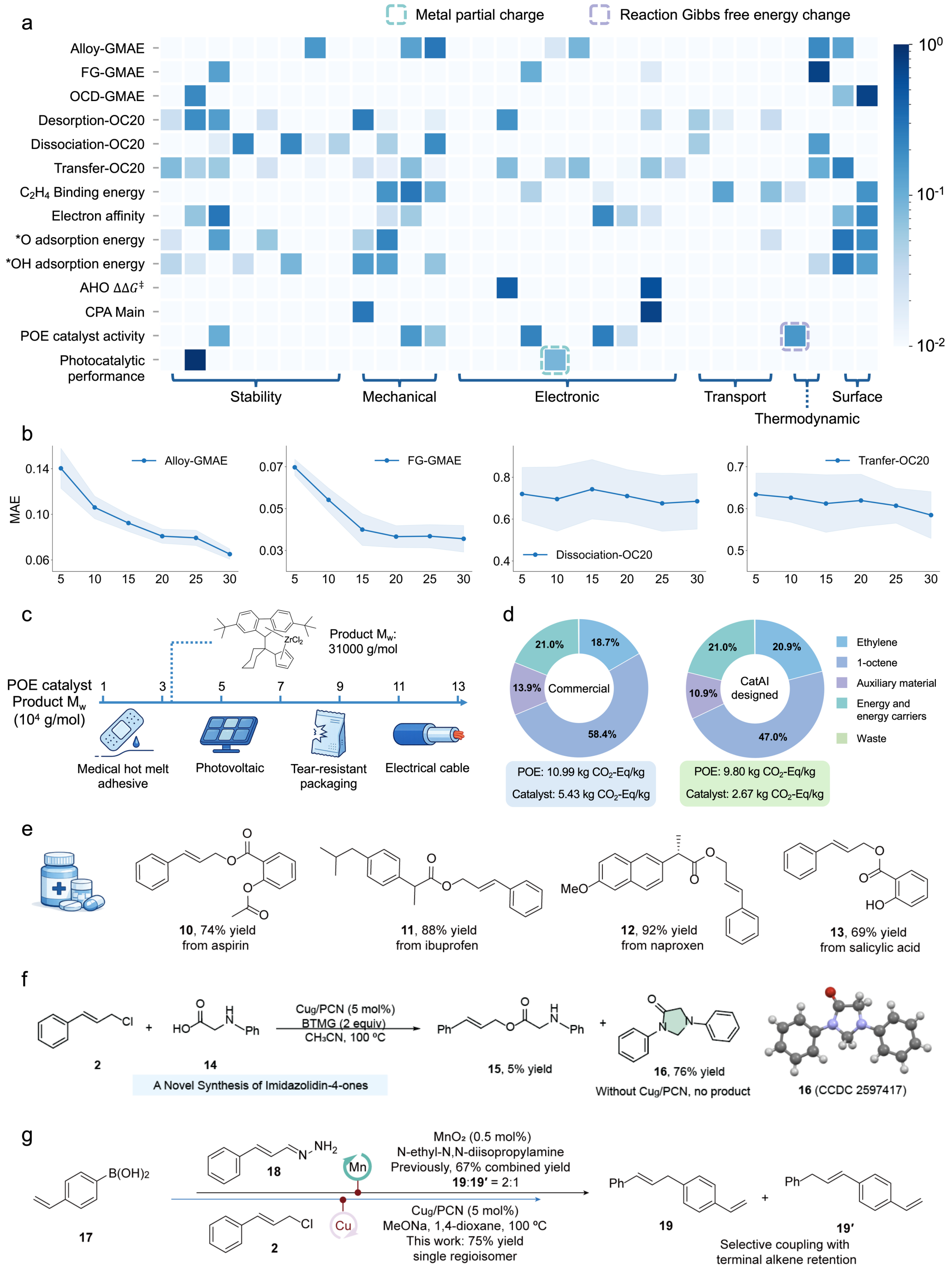

**Fig. 5 | CatAI as a composable knowledge infrastructure for continual catalysis discovery and downstream value.** (a) Task-conditioned module fusion weights visualized as normalized fusion weights across benchmark tasks and modules, showing structured reuse of physically meaningful priors. (b) Scaling of predictive accuracy with module library size, reported as MAE on representative adsorption and reaction-energy benchmarks as additional modules are made available for fusion. (c) Application context for the CatAI-designed catalyst **CPP-Zr** in high-value POE production. (d) Life-cycle assessment comparison between the CatAI-designed POE catalyst **CPP-Zr** and the commercial reference **CG-Ti** under matched process assumptions, indicating process-level benefits. (e) Late-stage functionalization of complex architectures. (f) A novel $Cu_g$/PCN-catalyzed decarboxylative cyclization route for the synthesis of imidazolidin-4-one **16**. (g) Comparison between the previous $MnO_2$-mediated route and the $Cu_g$/PCN approach for the synthesis of **19** through C($sp^3$)-C($sp^2$) bond formation.

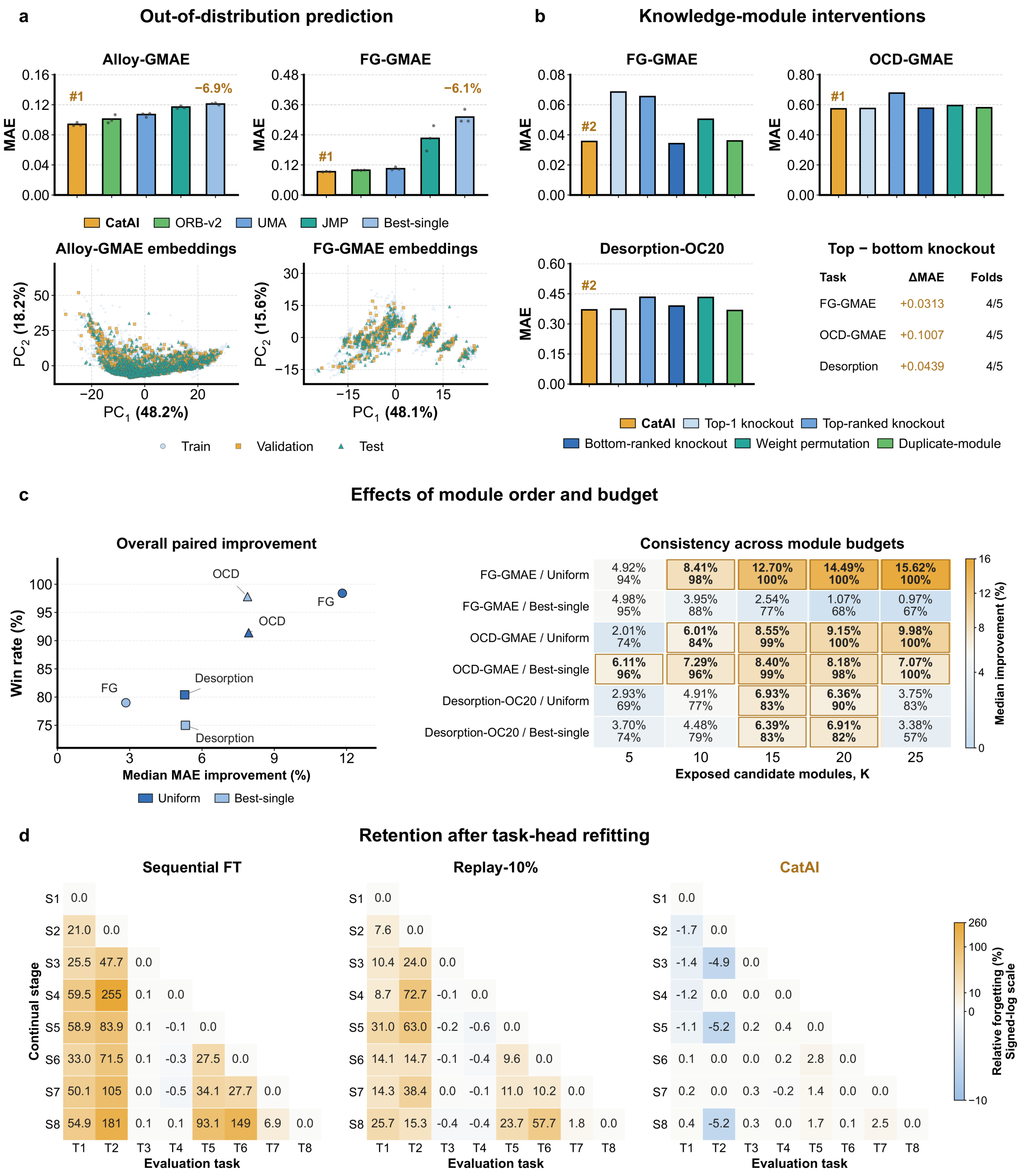

**Extended Data Fig. 1 | Additional evaluations of CatAI knowledge composition.** (a) OOD MAE for Alloy-GMAE and FG-GMAE against four baselines (three seeds). Percentages show reductions relative to the next-best mean. Lower panels show ID-seed0 embeddings after training-fitted standardization and PCA; axis percentages indicate explained training variance. These projections do not independently define OOD partitions. (b) Five-fold mean MAE for CatAI and five module interventions. The table reports paired top- versus bottom-ranked knockout differences and folds with higher MAE after top-ranked knockout. (c) Relative MAE improvement, $100 \times (1 - \mathrm{MAE}_{\mathrm{CatAI}} / \mathrm{MAE}_{\mathrm{baseline}})$, and win rate against Uniform or Best-single. Scatter plots summarize 500 selected cells per comparison; heatmaps report median improvements and win rates across 20 orders × five splits per budget. Gold outlines indicate ≥5% improvement and ≥75% wins. Orders were selected by test performance; this post-hoc analysis is not independent confirmation. (d) Relative forgetting after training-set task-head refitting and validation-based selection: $F[t,j] = 100 \times (\mathrm{MAE}[t,j] - \mathrm{MAE}[j,j]) / \mathrm{MAE}[j,j]$. Rows and columns denote stages and tasks, respectively. Sequential FT, Replay-10% and CatAI share a signed-log scale: blue, improvement; white, zero; gold, forgetting. Unobserved cells are blank. Error bars are omitted; seed/fold values and SDs are retained in the source data.

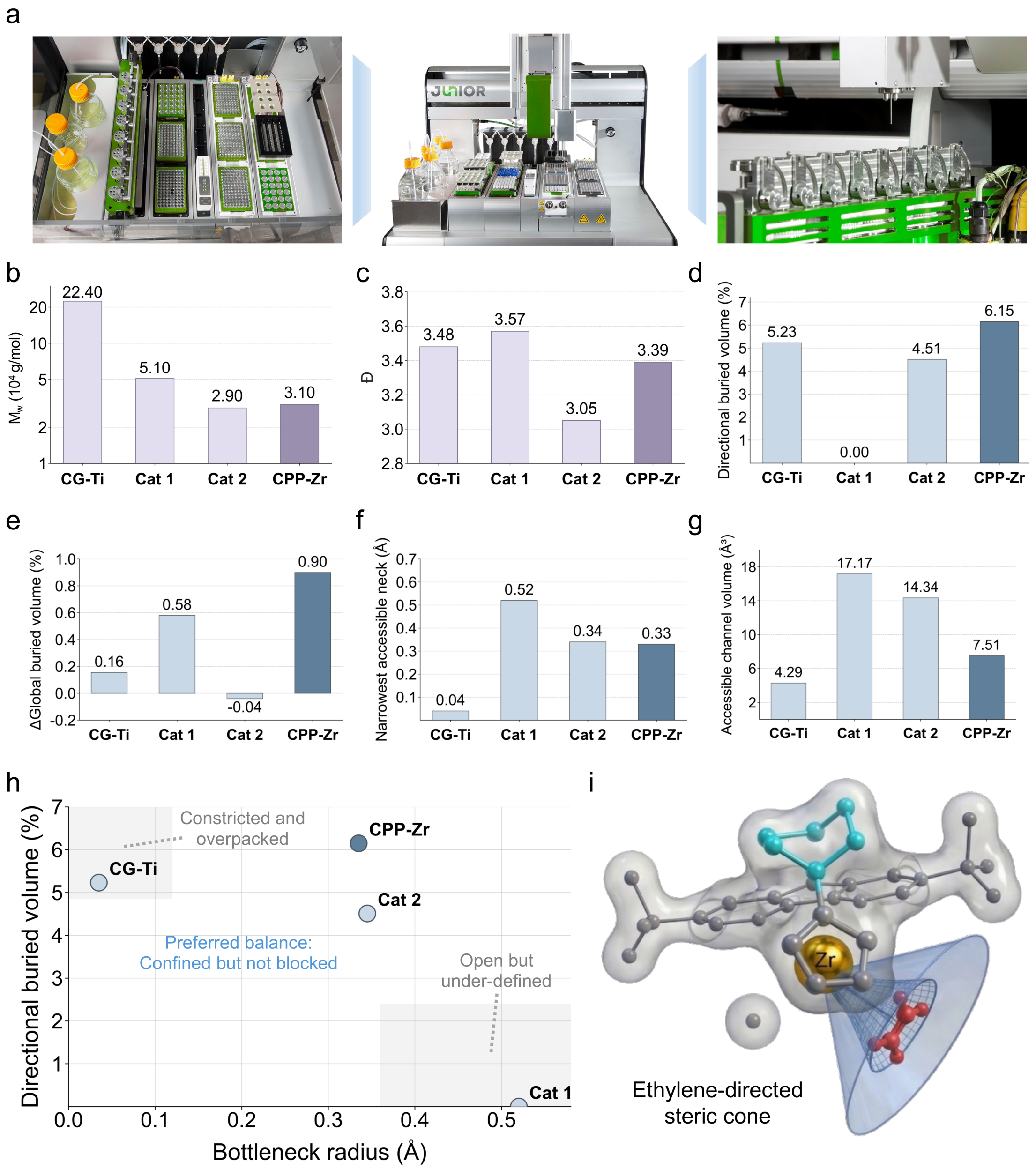

**Extended Data Fig. 2 | Experimental platform, polymer properties and steric analysis.** (a) Photographs of the automated polymerization evaluation platform. (b) Weight-average molecular weight ($M_w$) and (c) dispersity (Đ) of polymer produced by the commercial catalyst (**CG-Ti**) and three CatAI-recommended catalysts (**Cat 1**, **Cat 2** and **CPP-Zr**). (d) Directional buried volume projected along the ethylene-approach axis. Higher values indicate that ligand steric pressure is concentrated along the monomer insertion vector. (e) Change in global buried volume upon ethylene coordination, defined as the difference between the ethylene-coordinated state and the pre-coordination state. A larger positive shift signifies stronger coordination-induced tightening, whereby the ligand framework actively reorganizes around the incoming monomer to produce a more insertion-competent pocket. (f) Narrowest bottleneck radius of the catalytic channel, defining the most constricted cross-section through which the incoming monomer must pass. (g) Accessible channel volume surrounding the active site, reflecting the residual through-space capacity available for monomer approach and local chain reorganization. (h) Steric balance map correlating bottleneck radius with directional buried volume. (i) Three-dimensional visualization of the ethylene-directed steric cone and pocket map.

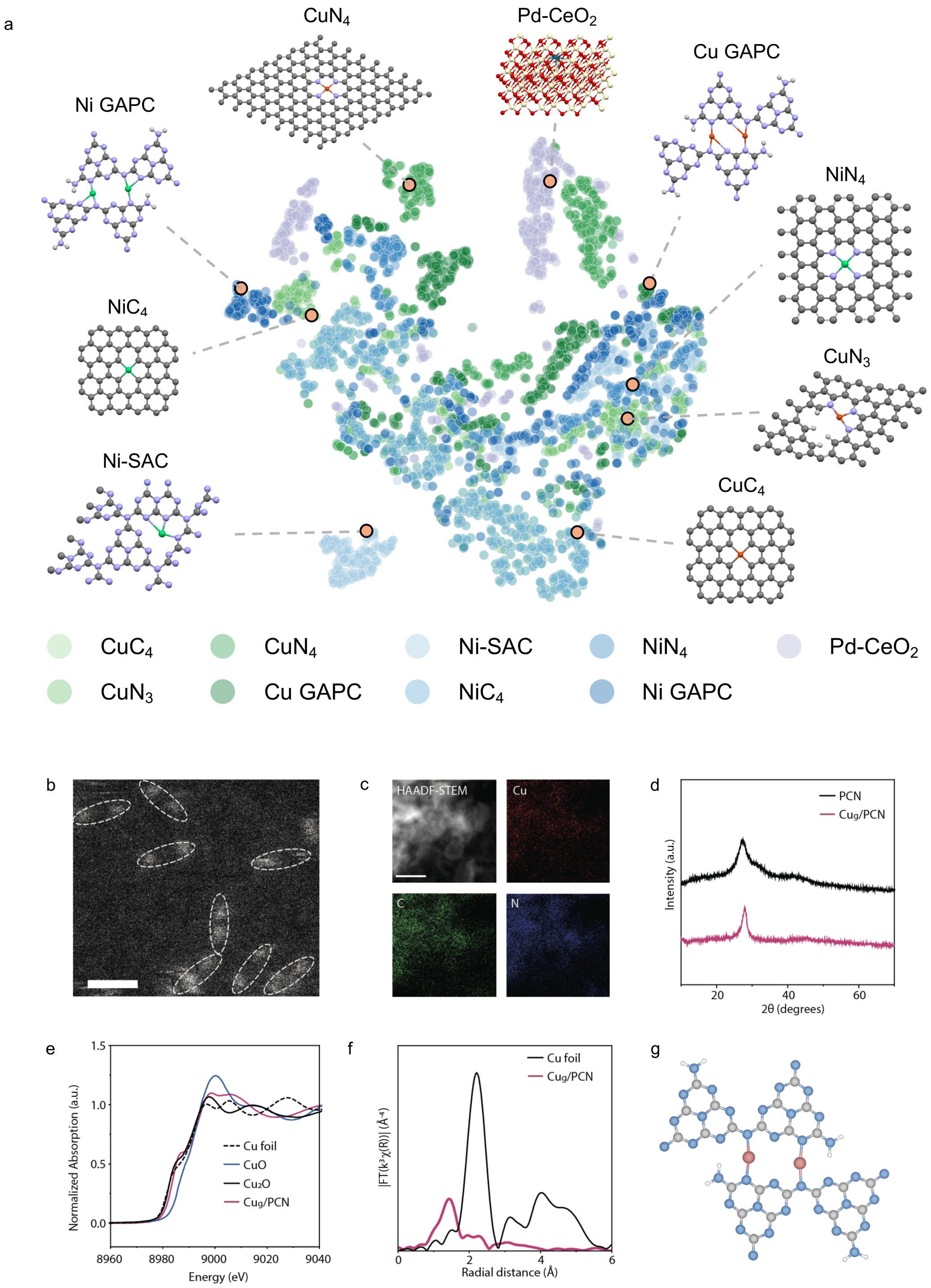

**Extended Data Fig. 3 | Catalyst design space and structural characterization of the Cu photocatalyst.** (a) Dimensionality-reduced visualization of the catalyst candidate library used in the active-learning campaign, spanning diverse metal centers, support materials and coordination motifs. (b–g) Structural characterization and proposed local configuration of **Cu GAPC** (**$Cu_g$/PCN**). (b) Atomic-resolution ADF-STEM image; geminal Cu structures are circled. Scale bar, 2 nm. (c) STEM-EELS mapping. Scale bar, 2 nm. (d) XRD patterns of pristine PCN and **$Cu_g$/PCN**. (e) Cu K-edge XANES spectra of **$Cu_g$/PCN** together with Cu foil, $Cu_2O$ and CuO references. (f) $k^3$-weighted Cu K-edge Fourier-transformed EXAFS spectra of **$Cu_g$/PCN** and Cu foil. (g) Proposed structural model. Atom colors: C, gray; N, blue; H, white; Cu, pink.